\documentclass[review,12pt]{elsarticle}

\usepackage{amssymb}
\usepackage{hyperref}

\usepackage{subcaption}

\journal{arXiv}

\begin{document}

\begin{frontmatter}



\title{ISeeU2: Visually Interpretable ICU mortality prediction using deep learning and free-text medical notes}


\author[AUT,UTB]{William Caicedo-Torres\corref{mycorrespondingauthor}}
\cortext[mycorrespondingauthor]{Corresponding author}
\ead{william.caicedo@aut.ac.nz}

\author[AUT]{Jairo Gutierrez}

\address[AUT]{Auckland University of Technology, Auckland - New Zealand}
\address[UTB]{Universidad Tecnol\'{o}gica de Bol\'{i}var, Cartagena - Colombia}

\begin{abstract}
Accurate mortality prediction allows Intensive Care Units (ICUs) to adequately benchmark clinical practice and identify patients with unexpected outcomes. Traditionally, simple statistical models have been used to assess patient death risk, many times with sub-optimal performance. On the other hand deep learning holds promise to positively impact clinical practice by leveraging medical data to assist diagnosis and prediction, including mortality prediction. However, as the question of whether powerful Deep Learning models attend correlations backed by sound medical knowledge when generating predictions remains open, additional interpretability tools are needed to foster trust and encourage the use of AI by clinicians. In this work we show a Deep Learning model trained on MIMIC-III to predict mortality using raw nursing notes, together with visual explanations for word importance. Our model reaches a ROC of 0.8629 ($\pm 0.0058$), outperforming the traditional SAPS-II score and providing enhanced interpretability when compared with similar Deep Learning approaches.

\end{abstract}




\end{frontmatter}


\section{Introduction}
Intensive Care Units (ICUs) are the last line of defense against critical conditions that require constant monitoring and advanced medical support. Their importance has been highlighted in recent times, when ICUs around the world have been overrun by the COVID-19 pandemic \cite{Grasselli2020, Emanuel2020}. It is in times like these when research into ways to adequately manage scarce critical care resources must be even more vigorously pursued, in order to offer additional tools that support medical decisions and allow for the effective benchmark of clinical practice.\\

The issue of mortality prediction in the ICU has been approached from a statistical standpoint by means of risk prediction models like APACHE, SAPS, MODS, among others \cite{Rapsang:2014aa}. These models use a set of physiological predictors, demographic factors, and the occurrence of certain chronic conditions, to estimate a score that serves as a proxy for the likelihood of death of ICU patients. Because of the relatively straightforward way of interpreting results, simple statistical approaches such as logistic regression are the go-to modeling techniques used to estimate mortality probability and the importance of the predictors involved. On the other hand, the simplicity of the models also mean that their limited expressiveness may not accurately represent the possibly non-linear dynamics of mortality prediction. Given this, high-capacity machine learning models might be useful to increase predictive performance. Concretely, the relevant literature shows that the use of deep learning models trained on physiological time-series data can outperform these previously mentioned statistical models \cite{PURUSHOTHAM2018, Caicedo-Torres2019}. \\

One of the advantages of deep learning over other techniques is its ability to use multiple modes of data to train predictive models. In the biomedical domain, health records, images, and time-series data, have been used for different tasks with success \cite{annurev-bioeng-071516-044442, Shickel2018}. This advantage is relevant for mortality prediction (and for many other clinical tasks as well), as a substantial amount of data is generated inside ICUs as free-text notes which can be used as input to create Natural Language Processing (NLP) predictive models. The nature of NLP poses some challenges for which deep learning is uniquely suited via its ability to deal with high-dimensional data and its elegant way to take temporal and spatial patterns into account. Some works have used deep neural networks and free-text to predict mortality \cite{Grnarova2016NeuralDE} and length of stay (among others), showing that there is interesting potential for this type of models.\\

On the other hand, a particularly important downside of deep learning is that, compared to the simpler logistic regression based models, feature importance is not as readily available. This in turn makes these models hard to interpret, as internally the model may transform the original input features to high-dimensional spaces via non-linear transformations, making it hard to establish the impact of each predictor on the predicted outcome. It has been documented that given their large predictive capacity, deep learning models can easily fit spurious correlations in the datasets used for their training, leading to potential diagnostic issues \cite{Cooper1997}. However some work has been done to interpret deep learning models in order to offer explanations intended to foster trust and further encourage their usage in the critical care setting. For instance, in our previous work we developed an interpretable deep learning mortality prediction model that uses physiological time-series data from the first 48 hours of patient ICU stay \cite{Caicedo-Torres2019}.\\

In this work, we present ISeeU2, a deep learning model that uses free-text medical notes from the first 48 hours of stay to predict patient mortality in the ICU. We use the MIMIC-III database \cite{Johnson:2016aa} to train a convolutional neural network (ConvNet) that is able to use raw nursing notes with minimal pre-processing to efficiently generate a prediction, and we couple the prediction of mortality with word importance and sentence importance visualizations, in a way that annotates the original medical note to show what parts of it are more predictive for death or survival, according to the model.

\subsection{Related work}

In the past some works have used deep learning to predict ICU mortality using free text. Grnarova et al \cite{Grnarova2016NeuralDE} proposed the use of a convolutional neural network for ICU mortality prediction using free-text medical notes from MIMIC-III. They used all medical notes from each patient stay to predict mortality, and trained their model using a custom loss function that included a cross-entropy term involving mortality prediction at the sentence as well, with promising results. Jo et al \cite{Jo2017CombiningLA} used a hybrid Latent Dirichlet Allocation (LDA) + Long Short Term Memory (LSTM) model for ICU mortality prediction trained on medical notes from MIMIC-III, in which the LSTM used the topic LDA features as input. Suchil et al \cite{Sushil2018} used stacked denoising autoencoders to create patient representations out of medical free-text notes, to be used for downstream tasks as mortality prediction. Si et al \cite{Si2019} proposed the use of a ConvNet for multitask prediction (mortality, length of stay), using all available patient medical notes up until time of discharge. Jin et al \cite{jin2018improving} proposed a multimodal neural network architecture and a Named Entity Recognition (NER) text pre-processing pipeline to predict in-hospital ICU mortality using all available types of free-text notes and a set of vital signs and lab results from the first 48 hours of patient stay, extracted from MIMIC-III.\\

Most of these works include some ad-hoc interpretability mechanism: Grnarova et al \cite{Grnarova2016NeuralDE} included a sentence-based mortality prediction target which is then used to score individual words according to their associated predicted mortality probability, Jo et al \cite{Jo2017CombiningLA} used LDA-computed weights to provide word importance, Suchil et al \cite{Sushil2018} used a gradient-based interpretability approach to compute the importance of words in the input notes.\\

Our work has key differences relative to those from the related literature. As opposed to \cite{Grnarova2016NeuralDE, Si2019}, we only use notes from the first 48 hours of patient stay instead of all notes available up until the time of discharge/death, and as opposed to cite{jin2018improving} we only use nursing notes and not the whole spectrum of notes available in MIMIC-III. Also from an interpretability standpoint we rely on a theoretically sound concept from coalitional game theory, known as the Shapley Value \cite{shapley:book1952}, instead of explainability heuristics. Finally our visualization approach puts emphasis on presenting results in a way that can be easily understood and it is useful for users.\\

The contributions of our work are summarized in the following:

\begin{itemize}

\item We present a model that is able to offer performance comparable to state of the art models that use physiological time series data, but only using raw nursing notes extracted from MIMIC-III.
\item Our approach only uses data from the first 48 hours of patient stay, instead of using data from the entirety of the stay. That makes our model more usable in a real setting as a benchmark tool.
\item Our approach to interpretability is based on a theoretically sound concept (the Shapley Value) and our visualizations provide a novel way to annotate clinical free-text notes to highlight the most informative parts for the prediction of mortality.

\end{itemize}

This paper is organized as follows: first we will show the overall distribution of our patient cohort dataset and its corresponding distribution of medical free-text notes. Then we will briefly describe our approach to interpretability using the Shapley Value, followed by a description of our convolutional architecture and experimental setting. Finally we will present and discuss our results and end with our conclusions and suggested future work.

\section{Methods and materials}
\subsection{Participants}
We used the Medical Information Mart for Intensive Care (MIMIC-III v1.4) to create a dataset for the training of our deep learning model. MIMIC-III contains ICU records including vitals, laboratory, therapeutical and radiology reports, representing more than a decade of data from patients admitted to the ICUs of the Beth Israel Deaconess Center in Boston, Massachusetts \cite{Johnson:2016aa}. The median age of adult patients (those with age $>$ 16y) is 65.8 years, and the median length of stay (LoS) for ICU patients is 2.1 days (Q1-Q3: 1.2-4.6) \cite{Johnson:2016aa}.\\

Our patient cohort was created using the following criteria: only stays longer than 48 hours were considered, in cases where patients were admitted multiple times to the ICU only the first admission was considered, and patients should have at least one free-text note recorded during their ICU stay. These criteria lead to a sample with $n=21415$. Table \ref{note-dist-table} shows the different types of medical notes included in our dataset together with their respective counts. \\ 

\begin{table}[h]
\centering

  \begin{tabular}{l  r r}
    \hline
    \textbf{Note Type} & \textbf{Count} & Percentage  \\ \hline
    Nursing/other & 83147 & 36.78\% \\ 
    Radiology & 61096 & 27.02\% \\ 
    Nursing & 43790 & 19.37\% \\ 
    Physician & 27789 & 12.30\% \\ 
    Respiratory & 5728 & 2.53\% \\ 
    General & 1775 & 0.78\% \\ 
    Nutrition & 1549 & 0.68\% \\ 
    Rehab Services & 521 & 0.23\%  \\ 
    Social Work & 501 & 0.22\% \\ 
    Case Management & 134 & 0.060\% \\ 
    Consult & 40 & 0.018\% \\ 
    Pharmacy & 12 & 0.0053\% \\  \hline
    Overall & 226082 & 100\% \\
    \hline
  \end{tabular}
  \caption{\label{note-dist-table} Distribution of free-text medical notes in our dataset.}
\end{table}

\begin{figure}[h!]
\centering
\includegraphics[scale=0.3]{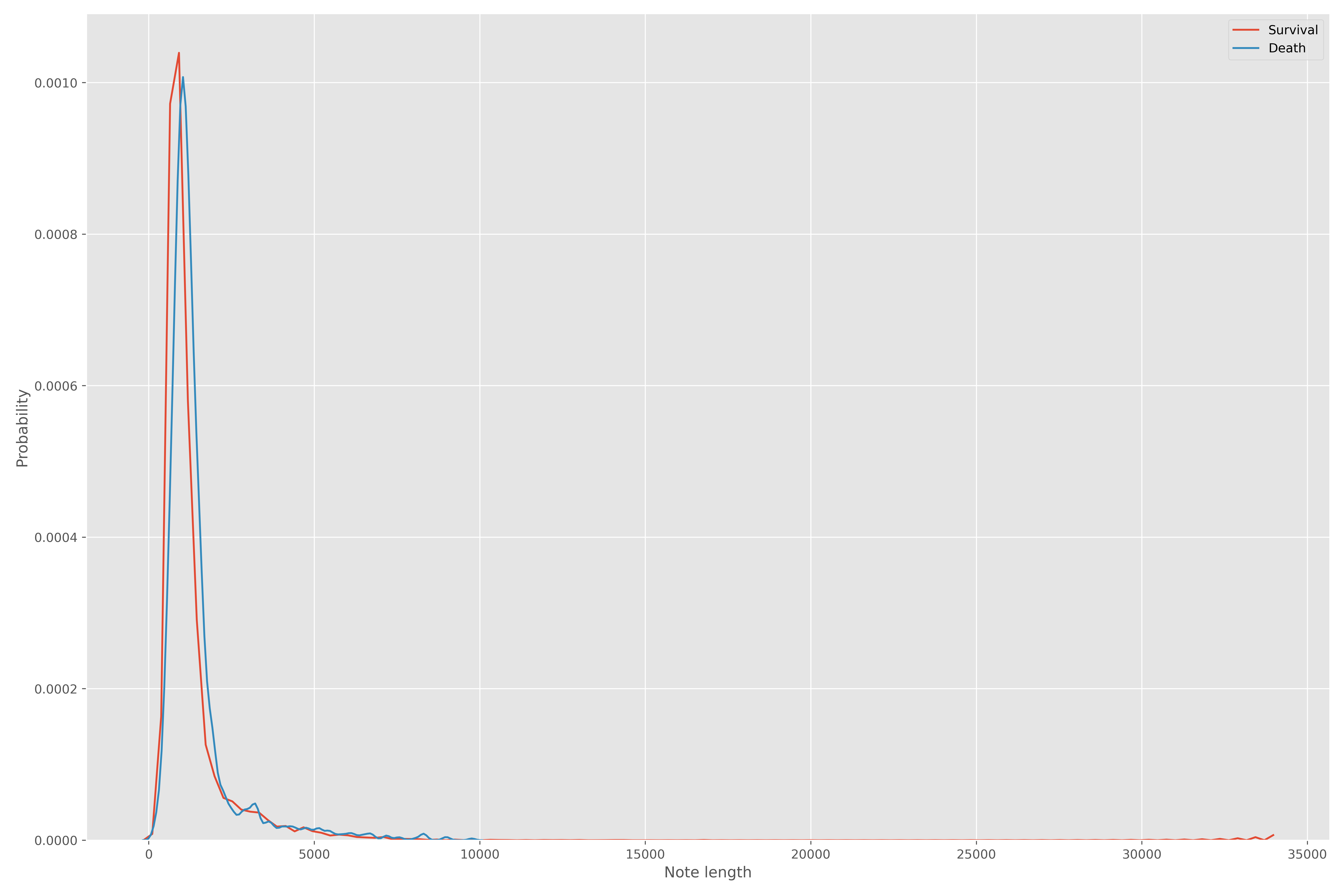}
\caption{Estimated nursing notes length distribution.}
\label{notes-length-box-plot}
\end{figure}

\begin{figure}[h!]
\centering
\includegraphics[scale=0.3]{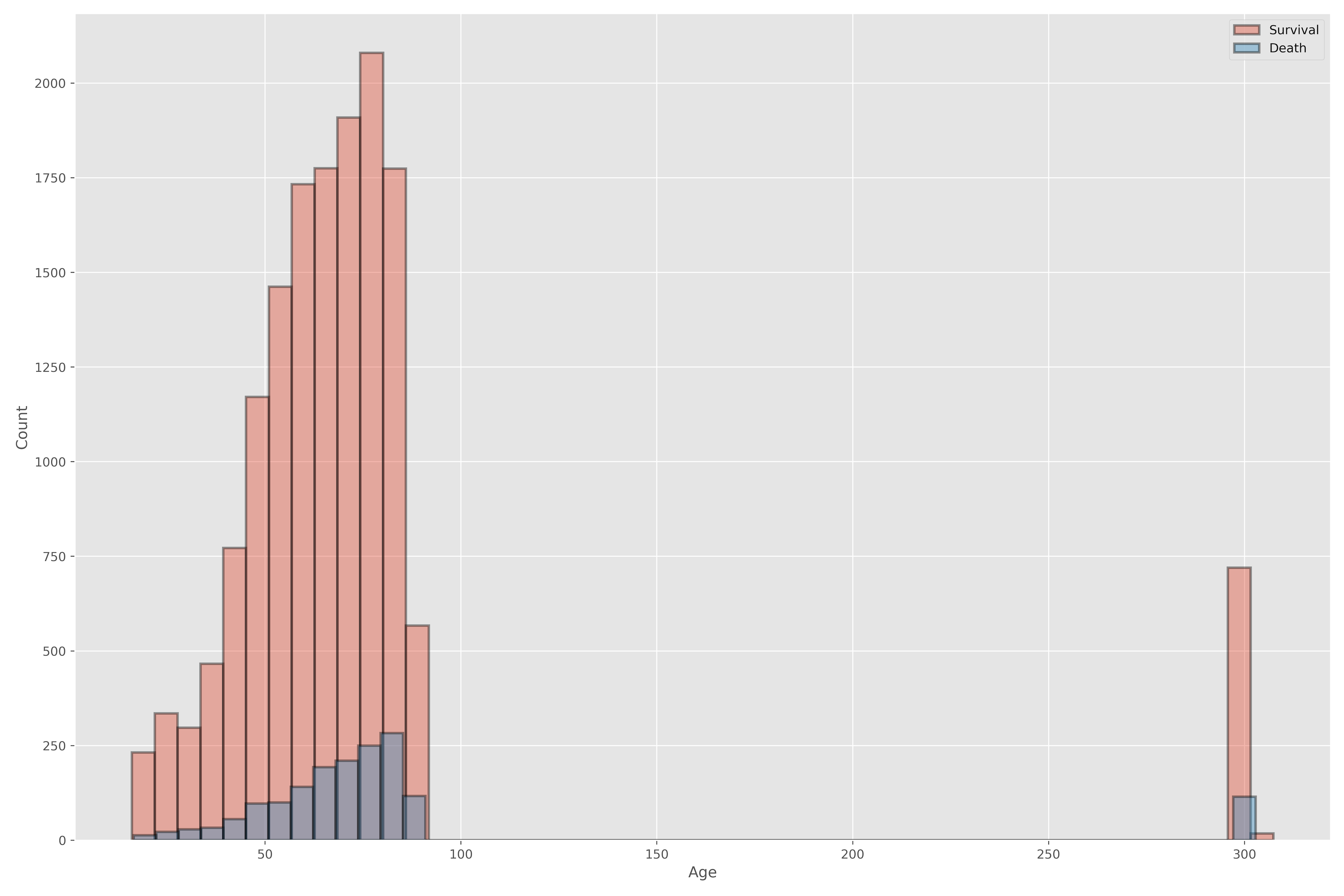}
\caption{Histogram of age distribution by outcome. As a result of privacy preserving measures, MIMIC-III shifts ages greater than 89 years (i.e. patients appear to be 300 years old). }
\label{age-hist}
\end{figure}

\begin{figure}[h!]
\centering
\includegraphics[scale=0.3]{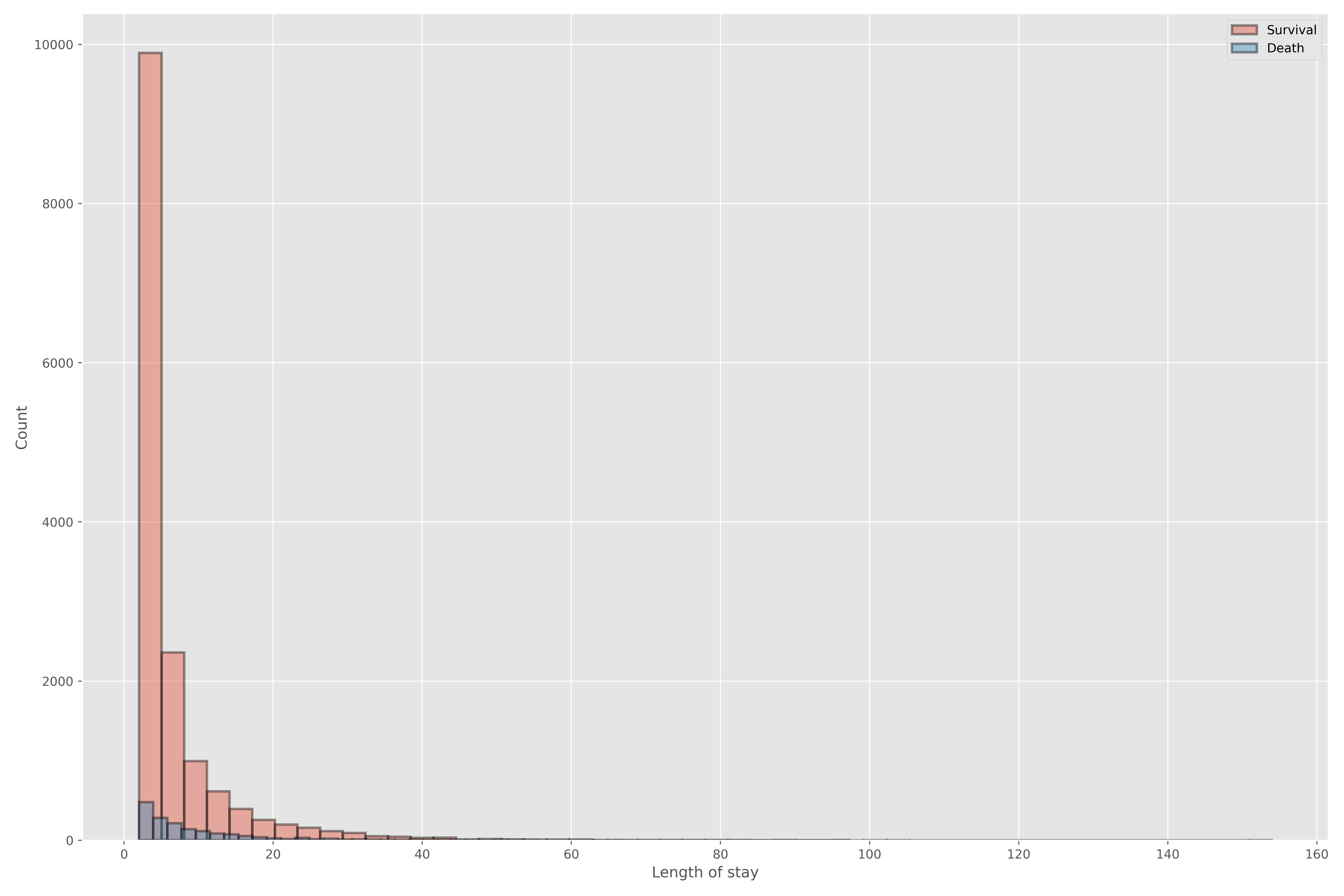}
\caption{Histogram of length of stay distribution by outcome.}
\label{los-hist}
\end{figure}

Given that a substantial number of patients in our dataset were missing more than one type of medical note, and that nursing and nursing/other types were the more prevalent ones, we decided to only include patients that had some type of nursing note available (nursing, nursing/other), with no regard to the note word count. This reduced our patient sample to $n=16970$, with 1659 recorded deaths (9.78\%) and 15311 patients that survived (90.22\%). The mean note length is 1252.59 words, with a standard deviation of 1087.48. Table \ref{note-length-dist} and figure \ref{notes-length-box-plot} show details about the distribution of notes' lengths. The median age of patients in our final sample is 67.2 years, and the median length of stay is 3.96 days (Q1-Q3:2.8-7.16). Figures \ref{age-hist} and \ref{los-hist} show the distribution of age and length of stay in our final sample.\\

\begin{table}[h]
\centering

\resizebox{\columnwidth}{!}{%
  \begin{tabular}{ l  r  r  r }
    \hline
    \textbf{Statistic} & \textbf{Positive class (survival)} & \textbf{Negative class (death)} & \textbf{Overall}\\ \hline
    Count & 15311 & 1659 &16970 \\ 
    Mean & 1233.3 & 1430.6 & 1252.6 \\ 
    Std & 1083 & 1112.4 &1087.5 \\ 
    Min & 34 & 144 & 34 \\ 
    Q1 & 711.0 & 890 & 724 \\ 
    Q2 & 934 & 1135 & 952 \\ 
    Q3 & 1286 & 1492 &1310 \\ 
    Max & 33771 & 9756 & 33771 \\  \hline
  \end{tabular}
  }
  \caption{\label{note-length-dist} Length distribution of nursing notes for our patient sample.}
\end{table}

\subsection{Deep learning model}
Our prediction model, called ISeeU2, is a convolutional neural network (ConvNet). ConvNets are a specialized neural network architecture that exploit the convolution operator and spatial pooling operations to detect local patterns and reduce input dimensionality to learn a representation that is useful for predictive purposes \cite{LeCun:1998aa}. ConvNets are extensively and primarily used for computer vision but have found application in Natural Language Processing as well, given their ability to deal with patterns occurring at different scales in sequential inputs \cite{Goodfellow-et-al-2016, Grnarova2016NeuralDE}. \\

\begin{figure}[h!]
\centering
\includegraphics[scale=0.4]{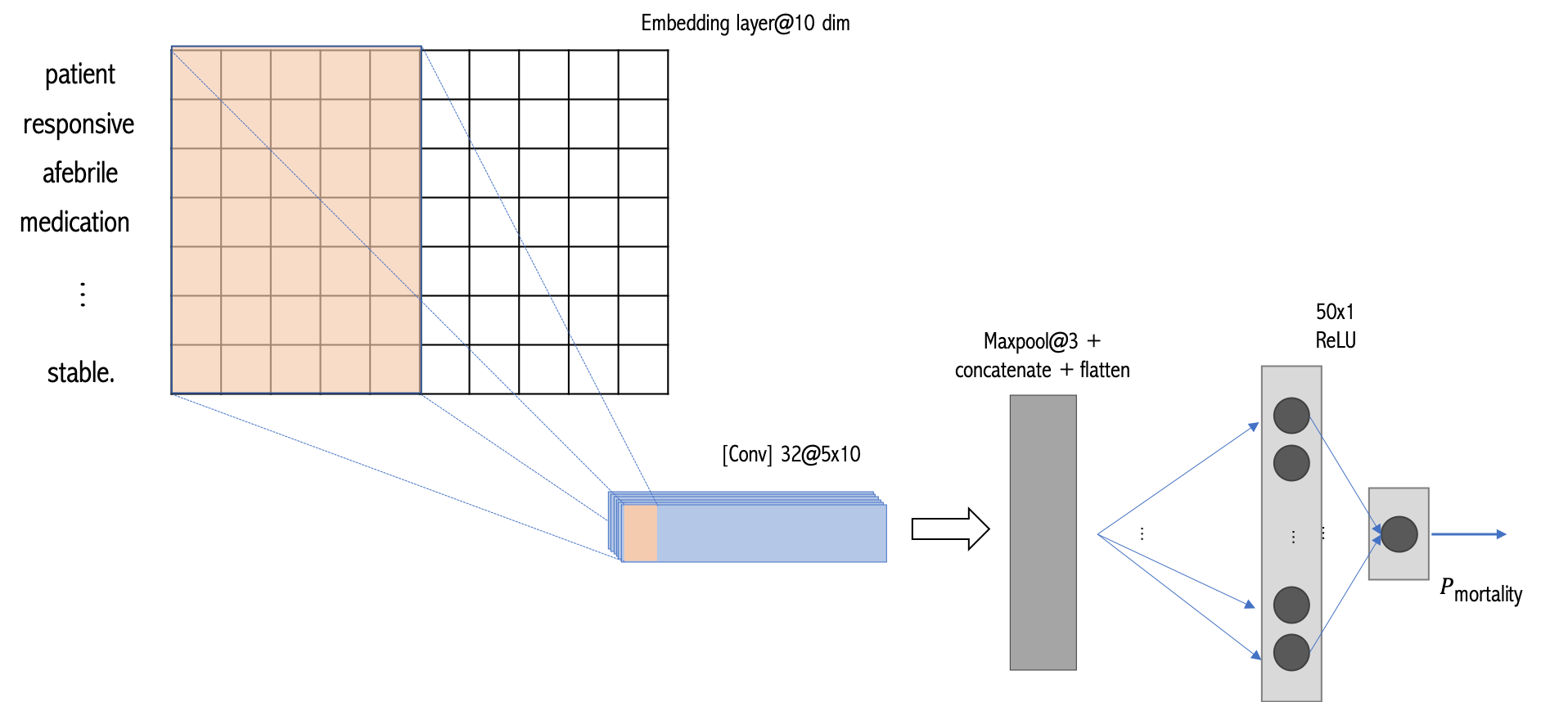}
\caption{Deep learning model architecture.}
\label{convnet-arch}
\end{figure}

The specific architecture of our model (figure \ref{convnet-arch}) includes a text embedding layer to convert a bag of words text representation into 10-dimensional dense word vectors. The output of the embedding layer is then fed to a convolutional layer with 32 channels and a kernel size of $5x10$ (stride 1), followed by ReLU activations and a max-pooling layer with a pool size of $1x3$ (stride 1). The obtained representation is then fed to a $50x1$ dense layer with ReLU activations connected to a one-neuron final layer with sigmoid activation, which computes the mortality probability.\\

One argument that is used routinely against deep learning is its reduced interpretability when compared to other modeling techniques such as logistic regression \cite{Cooper1997}. In order to overcome that potential limitation we use the Shapley Value in order to find how inputs affect the output of the model, hence gaining insight about which kinds of note fragments are more correlated with negative outcomes. The Shapley Value is a concept from coalitional game theory that formalizes the contribution of individual players towards the attainment of a goal as part of a team. The Shapley value captures the marginal importance of each player when its role is analyzed across all possible subsets of players from the original coalition. According to Shapley \cite{shapley:book1952}, given a coalitional form game $\langle N, v\rangle$, with a finite set of players $N$ of size $n$ and a function $ v:2^N \rightarrow \mathbb{R}$ that describes the total worth of the coalition, the marginal importance of player $i$ can be expressed as

\begin{equation} \label{eq:shapley}
Sh_i(v)=\sum_{S \subseteq N\setminus\{i\},s= |S|}\frac{(n-s-1)!s!}{n!}(v(S \cup \{i\})-v(S))
\end{equation} 

The summation is taken over all possible subsets $S \subseteq N$ that don't include player $i$, and each of its terms captures the effect of player $i$ on the reward attained by each subset, $v(S \cup \{i\})-v(S)$. \\

Strumbelj et al \cite{Strumbelj2010} have shown in their work that it is possible to apply the Shapley Value to the problem of feature importance quantification, if inputs are considered players in a coalition, and the predicted value is akin to the attained reward. In this way the Shapley Value becomes very useful, as it takes into account the interaction between features, in a way other methods like tree-based feature importance or simple input occlusion cannot. \\

A downside of using the Shapley Value for model interpretability is that equation \ref{eq:shapley} has combinatorial cost, and that's why using it may be unfeasible for practical purposes. In order to get around this limitation we resort to a fast approximate algorithm, DeepLIFT \cite{DBLP:journals/corr/ShrikumarGK17}, to compute approximate Shapley Values in feasible time \cite{Lundberg2017}. DeepLIFT is an algorithm specifically designed to compute feature importance in feed-forward neural networks. DeepLIFT overcomes the issues associated with competing methods such as Layerwise Relevance Propagation \cite{DBLP:journals/corr/ShrikumarGK17}, and gradient-based attribution \cite{SimonyanVZ13, DBLP:journals/corr/SpringenbergDBR14}, i.e. saturation, overlooking negative contributions, and gradient discontinuities \cite{DBLP:journals/corr/ShrikumarGK17}. DeepLIFT computes feature importance by comparing the network output to a reference output obtained by feeding the network with a designated input. The difference in outputs is back-propagated through the different layers of the network until the input layer is reached and feature importances are fully computed. A more detailed treatment of DeepLIFT in the context of interpreting deep learning models for critical care prognosis can be found in \cite{Caicedo-Torres2019}.\\ 

\section{Results}

Our ConvNet was built using Tensorflow \cite{Abadi2015}. Since our dataset is highly unbalanced (negative outcomes represent just 9.78\% of training examples), we used a weighted logarithmic loss assigning more importance to the positive class, i.e. patients that died in the ICU. We used 5-fold cross-validation to assess the model performance and place a confidence estimate on it. We did not perform any substantial hyperparameter optimization other than conservatively varying the number of channels of the convolutional layer and the number of neurons of the first fully connected layer of the network. Our choice of optimizer was Adam \cite{2014arXiv1412.6980K} with default Tensorflow-provided parameters. Our model was trained for three epochs per training fold, and we kept the lowest loss model of each run.\\

\subsection{Text pre-processing} One of our goals is to show a deep learning model that needs little to no input pre-processing in order for it to be as widely applicable as possible. Keeping with that we used the NLTK library \cite{Loper2002} to remove English stop-words and the Tensorflow.keras default tokenizer to vectorize the text notes, keeping the 100k most frequent words; and no further pre-processing was attempted. The tokenizer was fitted only on the training folds to avoid data leakage. Finally, we set the maximum note length to 500, so notes with a larger word count were truncated at the beginning and those with a smaller word count were padded at the beginning with zeroes. \\

\subsection{Model performance and comparison with baselines}
\begin{figure}[hbt!]
  \centering
    \includegraphics[width=\textwidth]{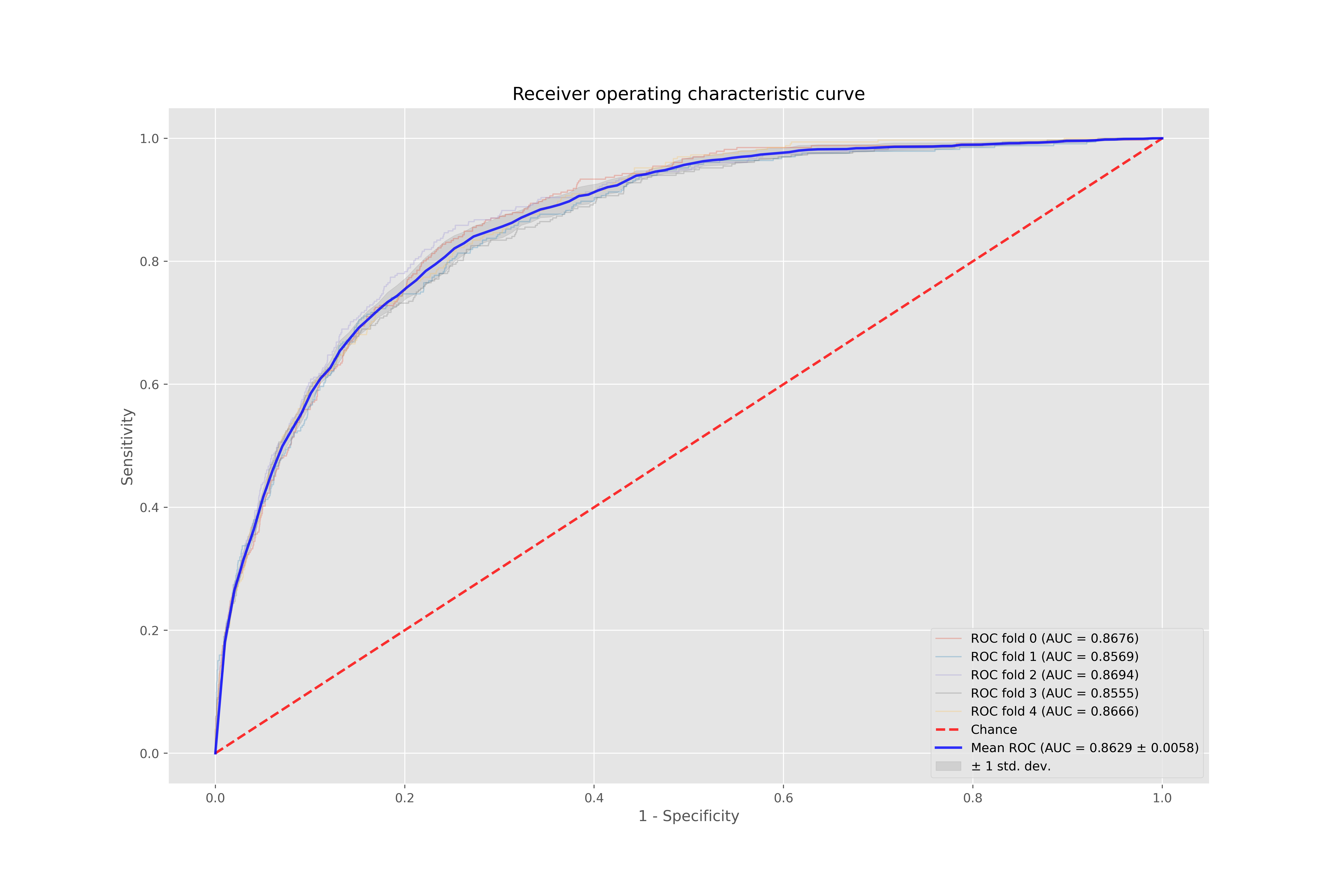}
    \caption{\label{validation_roc} ConvNet 5-fold cross validated ROC AUC. }
\end{figure}

Using this configuration we obtained a 5-fold cross validation Receiver Operating Characteristic Area Under the Curve (ROC AUC) of 0.8629 ($\pm 0.0058$) as seen in figure \ref{validation_roc}. Using a 0.5 decision threshold, the model reaches 72\% sensitivity at 83\% specificity. We also provide some baseline models to compare with our proposed model to better assess its performance. Concretely, we have included results for a traditionally used mortality risk score and a recurrent neural network.

\begin{figure}[hbt!]
  
  \centering
    \includegraphics[width=\textwidth]{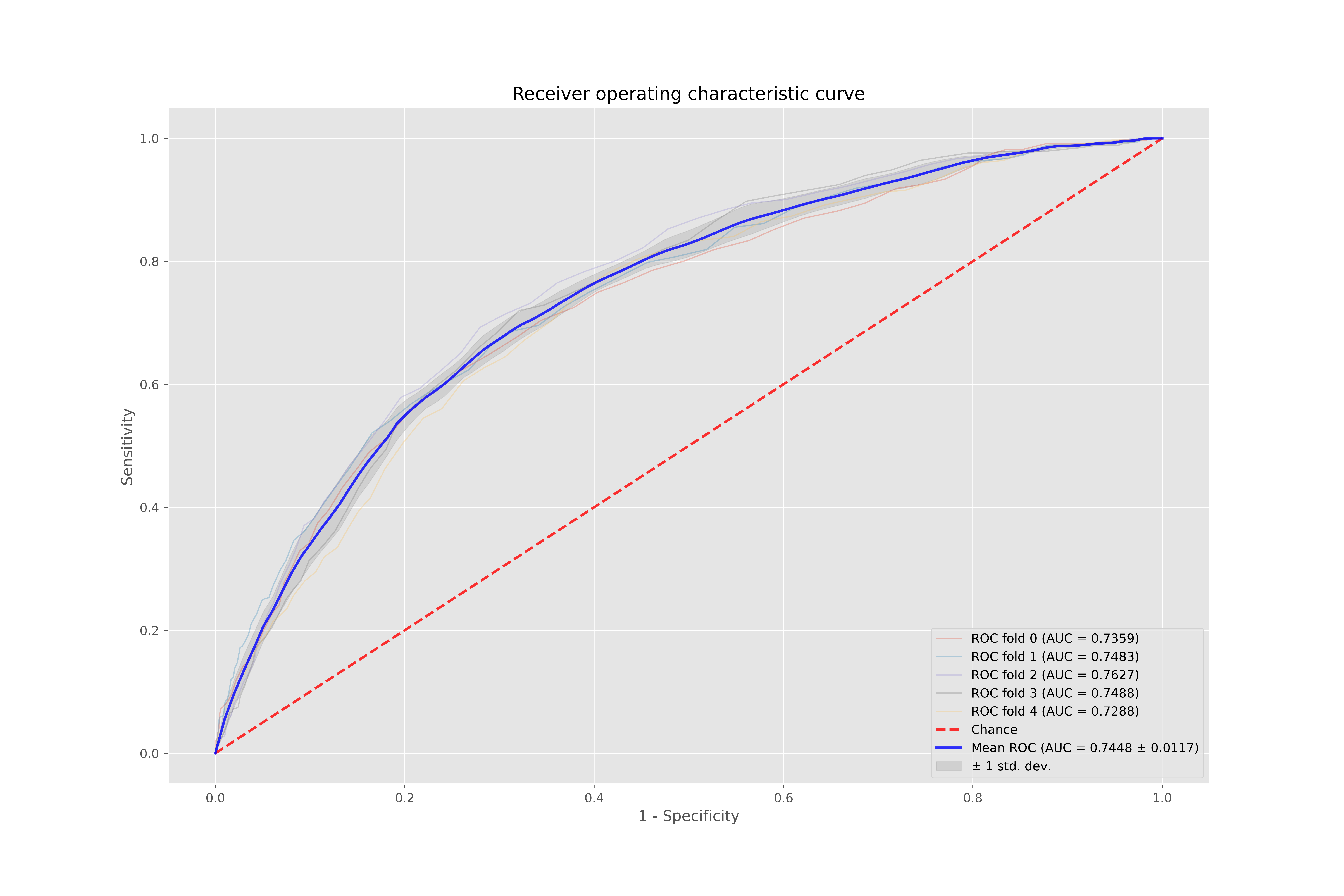}
    \caption{\label{SAPS-II_validation_roc} SAPS-II model 5-fold cross validated ROC AUC. }
\end{figure}

\begin{figure}[hbt!]
  
  \centering
    \includegraphics[width=\textwidth]{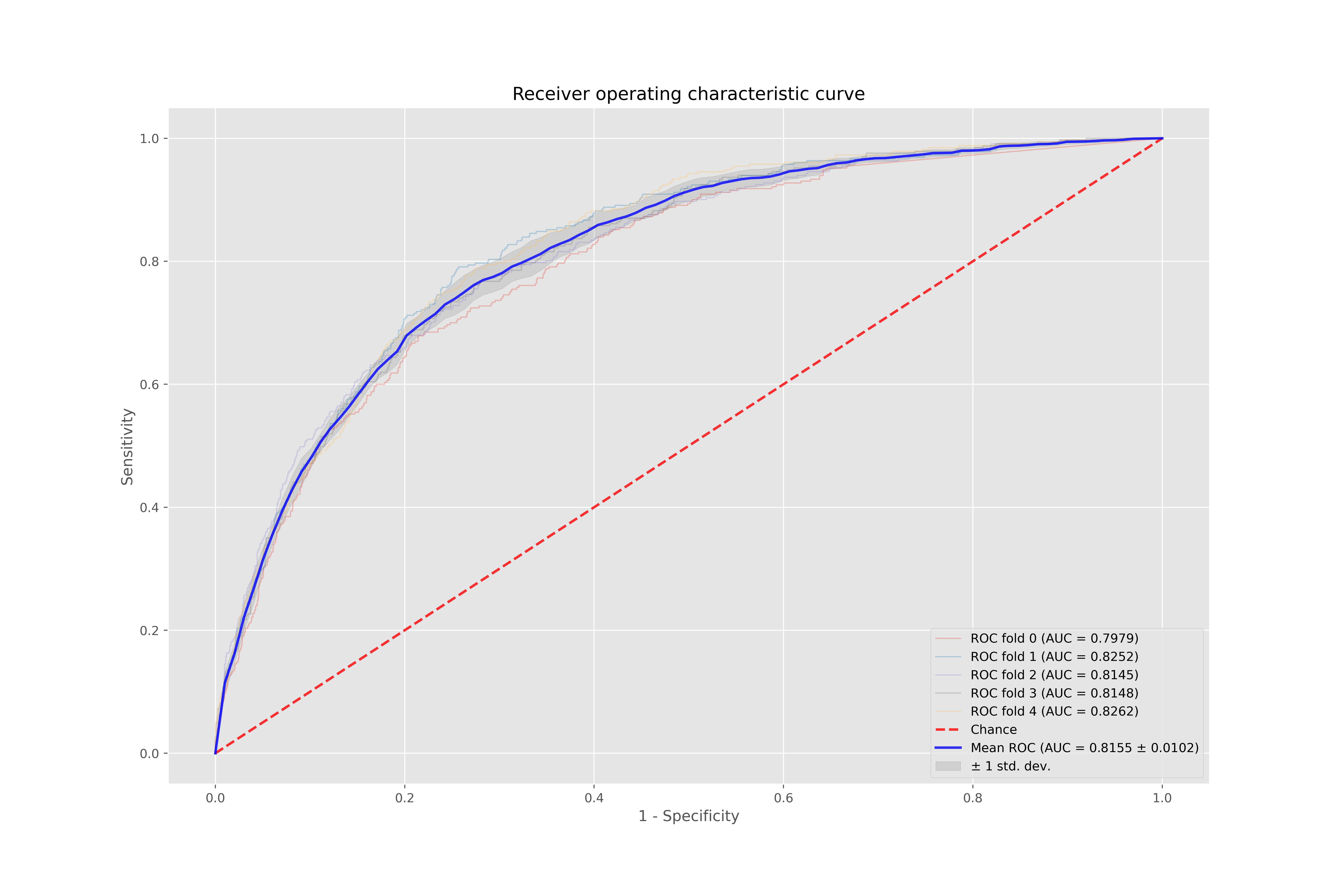}
    \caption{\label{validation_roc_24h} ConvNet (24h) 5-fold cross validated ROC AUC. }
\end{figure}

\paragraph{SAPS-II}
As baseline, we used a well-established ICU mortality risk score, SAPS-II \cite{Gall1993}. SAPS-II uses data from the first 24 hours of ICU stay to calculate a numerical score, which in turn is converted into a mortality probability. In order to compare our approach with SAPS-II predictions and performance, we trained our convolutional architecture using nursing notes from the first 24 hours only while keeping training parameters the same. We used the SAPS-II implementation provided by the authors of the MIMIC-III code repository \cite{Johnson2018}. The 24 hour version of our model obtained a 0.8155 ($\pm 0.0102$) ROC AUC 5-fold cross-validation score, against 0.7448 ($\pm 0.0117$) for the SAPS-II model. Figures \ref{SAPS-II_validation_roc} and \ref{validation_roc_24h} show the corresponding ROC plots for the two models.

\begin{figure}[hbt!]
  
  \centering
    \includegraphics[width=\textwidth]{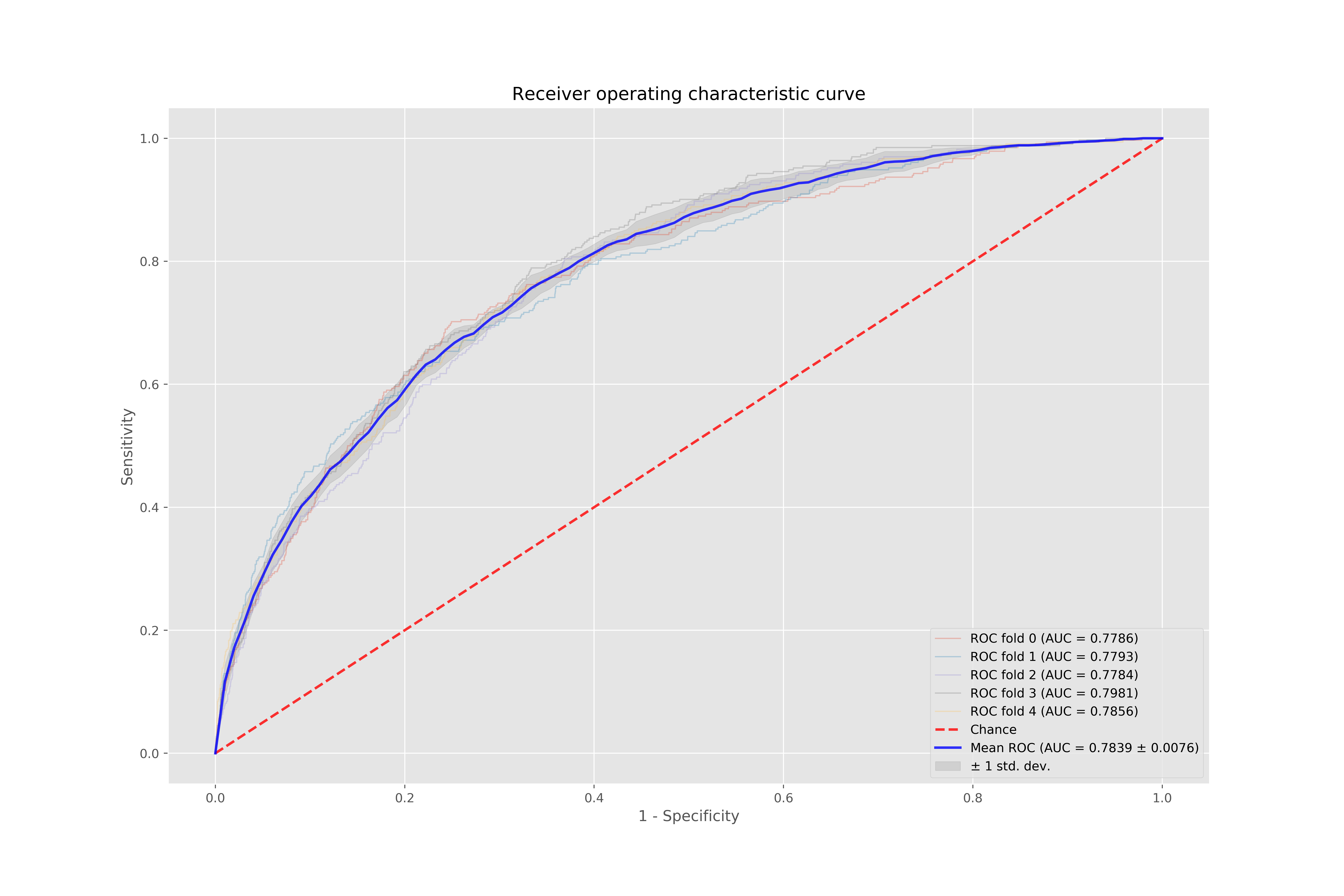}
    \caption{\label{lstm-roc} Deep LSTM 5-fold cross validated ROC AUC. }
\end{figure}

\paragraph{LSTM}
Our second baseline is a recurrent neural network based on the Long Short Term Memory (LSTM). LSTM is a neural network model designed to handle sequential input data with temporal dependencies \cite{Hochreiter:1997:LSM:1246443.1246450}, and it has been used extensively in Natural Language Processing tasks. We trained a deep neural network with a bidirectional LSTM layer with 100 units, followed by an extra 100-unit LSTM layer, a 50-unit dense layer ReLU activation, and a final sigmoid layer. As it was the case for our original convolutional model, an embedding layer was used to create 10-dimensional dense vectors to feed the initial layer of the LSTM and the same text preprocessing pipeline was used (save for a now 1000-word maximum note length). Finally dropout with probability $0.5$ was applied to control overfitting. With this particular architecture we were able to obtain a  0.7839 ($\pm 0.0076$) ROC AUC 5-fold cross-validation score (Figure \ref{lstm-roc}).

\subsection{Model interpretability}
\begin{figure}[!h]
  
  \centering
  \begin{subfigure}{0.8\textwidth}
  \centering
    \includegraphics[width=0.8\textwidth]{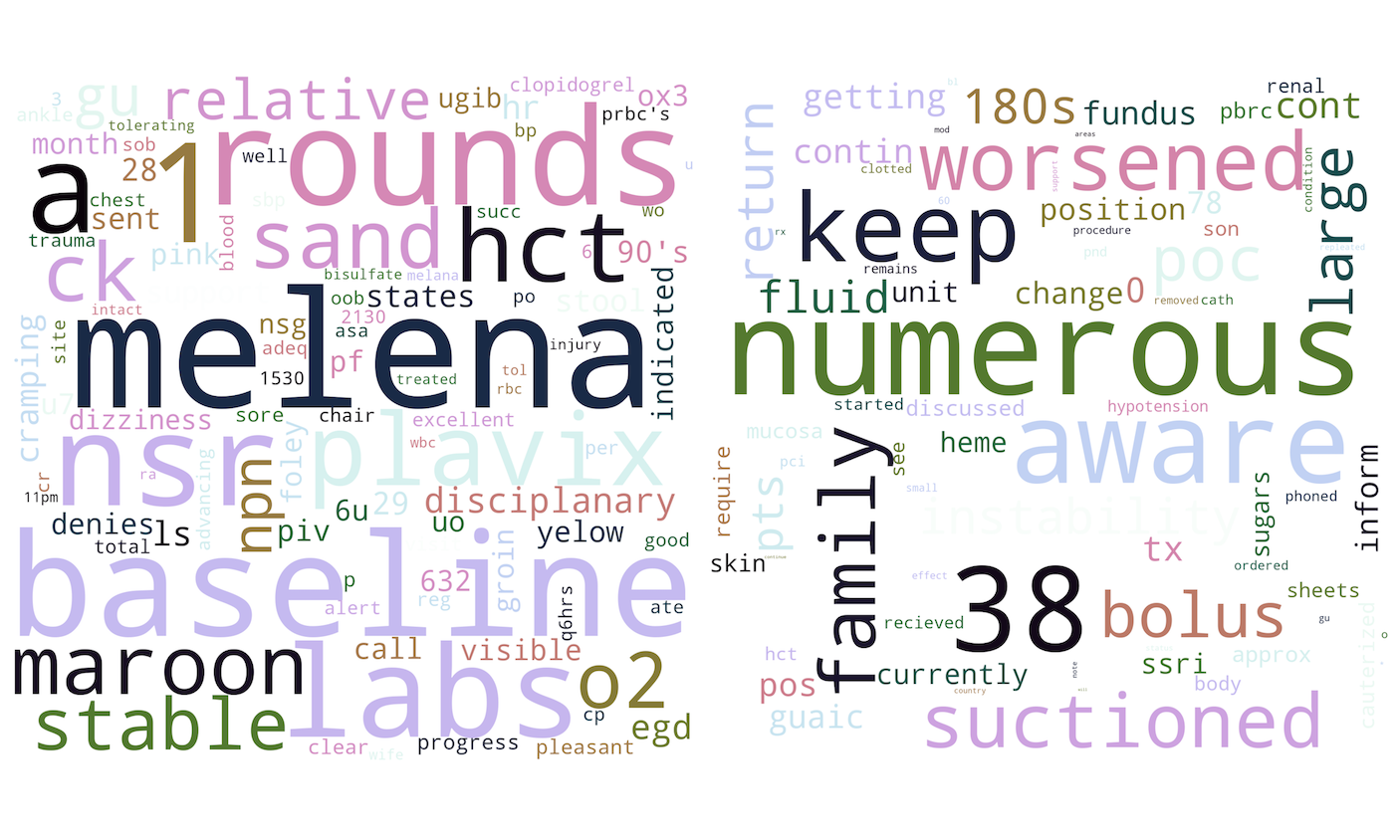}
\end{subfigure}
  \begin{subfigure}{\textwidth}
  \centering
    \includegraphics[width=0.8\textwidth]{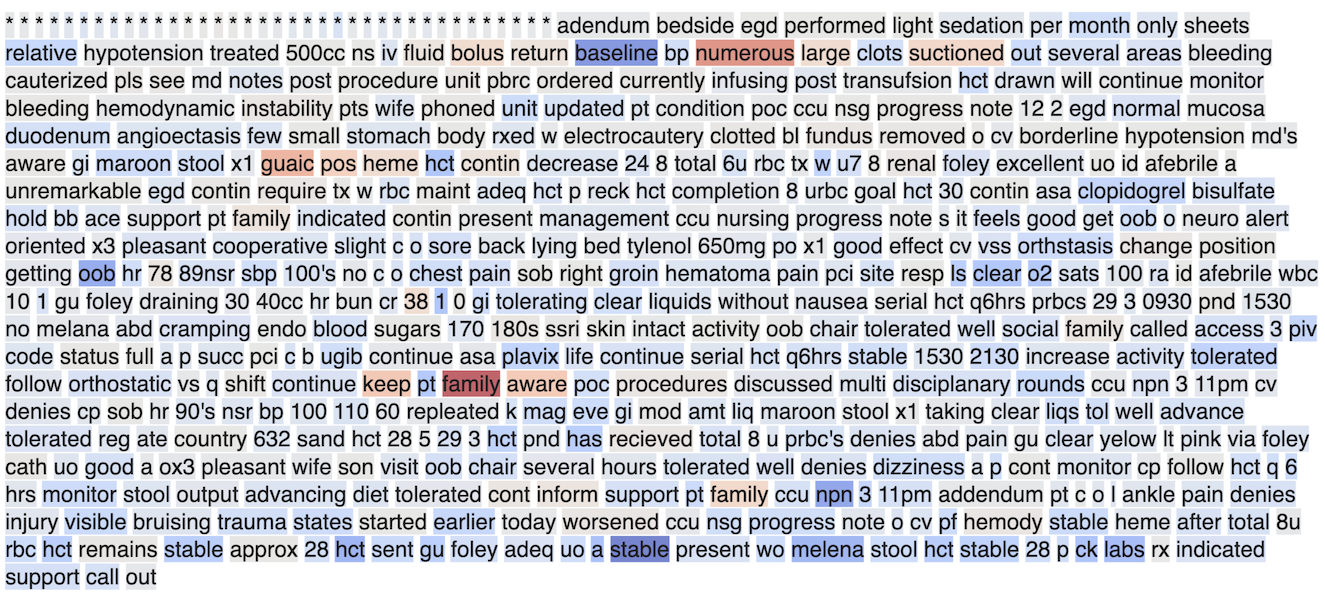}
    \end{subfigure}
    
    \caption{\label{train-text-heatmap} Top: Word clouds generated for one specific patient in the training set show the words deemed as most important for both survival (left) and death (right) prediction. Bottom: Text heatmap showing words, their importance and their context in sentences, generated for one specific patient in the training set. Red color denotes evidence for death, and blue color represents evidence for survival. Words with a gray background are not considered important for the prediction task by the network. Padding characters are represented by asterisks.}
\end{figure}

\begin{figure}[!h]
\centering

\begin{subfigure}{.40\textwidth}
\includegraphics[width=\textwidth]{pics/training-set-note-heatmap.png}
\end{subfigure}
\begin{subfigure}{.40\textwidth}
\includegraphics[width=\textwidth]{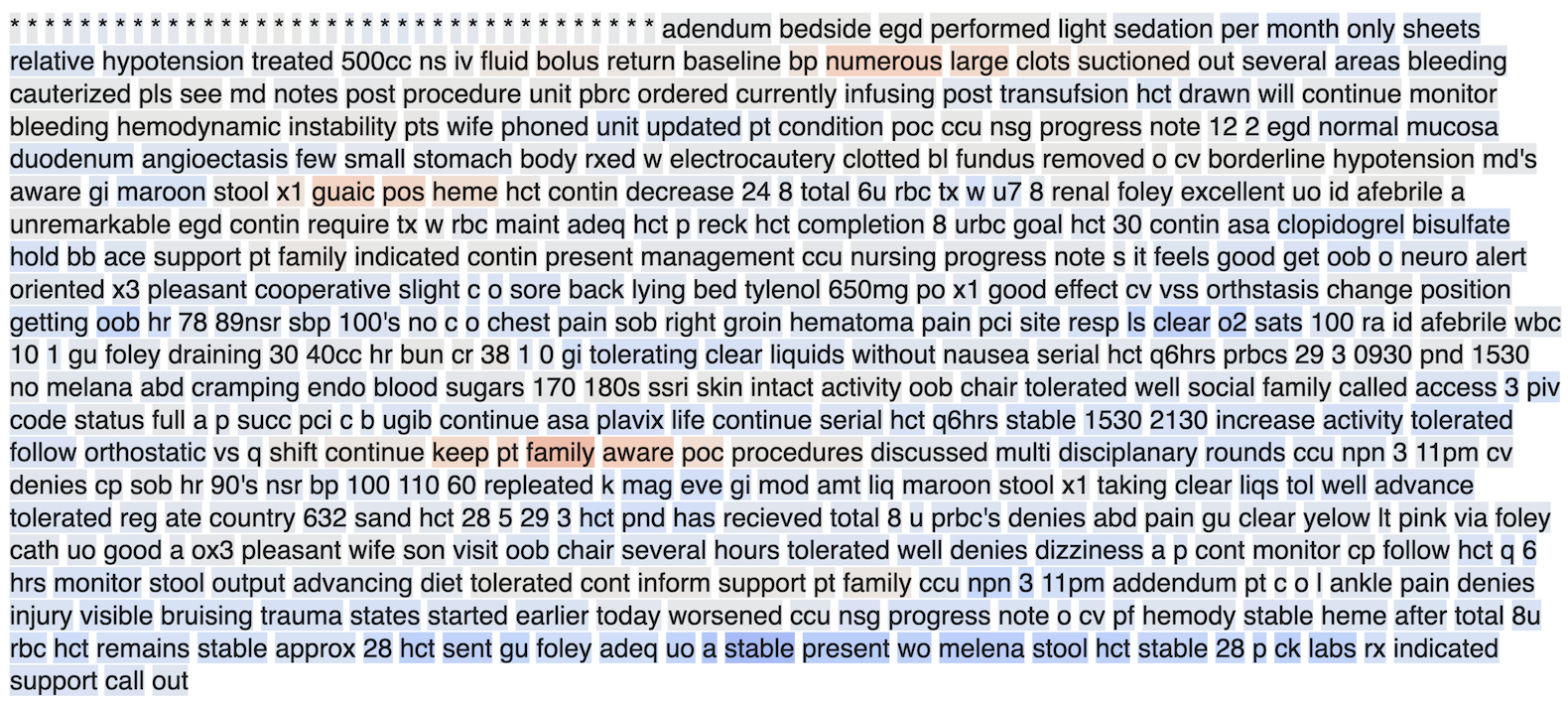}
\end{subfigure}

\begin{subfigure}{.40\textwidth}
\includegraphics[width=\textwidth]{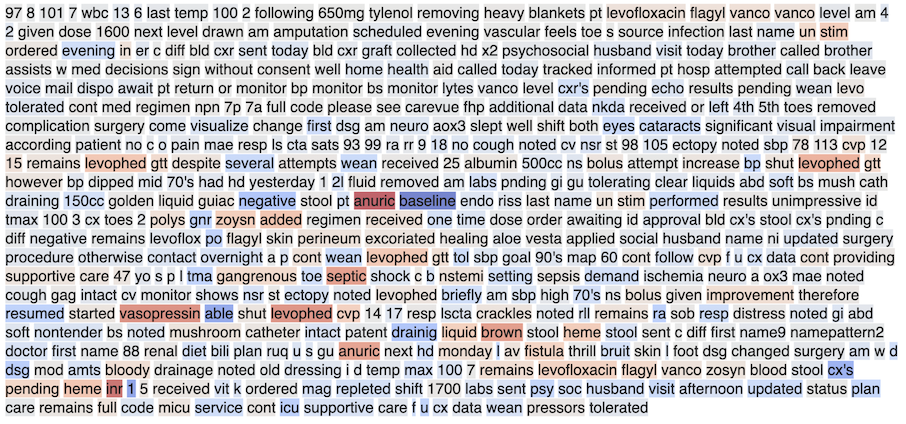}
\end{subfigure}
\begin{subfigure}{.40\textwidth}
\includegraphics[width=\textwidth]{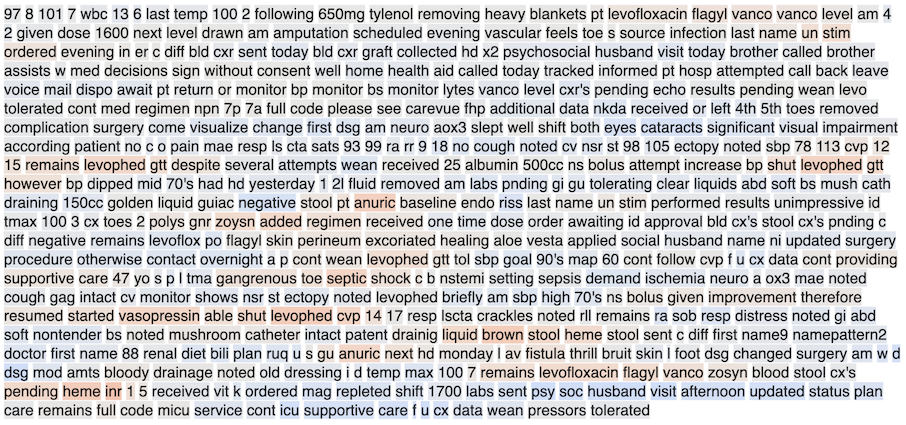}
\end{subfigure}

\caption{Text heatmaps with (right) and without (left) convolution smoothing. Bottom row corresponds to a nursing note from the validation set.}
\label{validation-text-heatmap}
\end{figure}

Using the DeepLIFT implementation provided by \cite{Lundberg2017} which works appropriately with Tensorflow 2 models, we calculated word importances for our model, using the empirical mean of the input embedding vectors as reference value. Using these values we designed and built visualizations to show the importance of each word in the original nursing note used as input. Our visualizations constitute a form of \textit{post hoc interpretability} \cite{DBLP:journals/corr/Lipton16a} insofar as they try to convey how the model regards the inputs in terms of their impact on the predicted probability of death, without having to explain the internal mechanisms of our neural network, nor sacrificing predictive performance. We have selected some examples at random from both the training set and the validation set of the last cross-validation run to show the behavior of the model and the way it regards certain words in the input notes. Our proposed visualizations include word clouds and text heatmaps (Figures \ref{train-text-heatmap} and \ref{validation-text-heatmap} ). \\

Word clouds are an interesting way to visualize words and their importance at the same time, but they don't capture the context in which words live, potentially leading to erroneous interpretations. For example, the survival word cloud in Figure \ref{train-text-heatmap} shows \textit{melena} as associated with survival, which is not readily understandable. However, when the word cloud is combined with the note heatmap, the reason becomes apparent, given the context of the word (\textit{stable present wo melena stool}). We also observe that certain phrases and words are flagged intuitively, e.g. \textit{guaic pos heme}, and also the fact that for this particular patient occurrences of Plavix/Clopidogrel in the note are flagged as evidence for survival. There are other instances in which results are not intuitive and may point to statistical flukes rather than strong causal features. As an example we can point to the phrases \textit{return baseline bp numerous large clots suctioned}, and \textit{continue keep pt family aware}, in which the words \textit{clot} and \textit{pt} seem to be flagged incorrectly. 

\paragraph{Annotation smoothing} In order to help ameliorate the sharp changes and inconsistencies observed at the sentence level we used a convolution filter to take into account the effect of the Shapley Values of all words in a particular sentence when generating the heatmap annotations, and provide a smoother and more intuitive result. Note that this is approximated since we are intent on using a basic pre-processing pipeline,, without any advanced capabilities (i.e. sentence segmentation). A $5 \times 1$ convolution filter $[0.1, 0.2, 0.4, 0.2, 0.1]$ allows us to spread out the feature importance of individual words and to fade out weak importances that are due possibly to noise, while still keeping the most salient features. Figure \ref{validation-text-heatmap} show our previous training set and a new validation set note with and without the convolution filter applied.

\paragraph{Note length and mortality probability} High capacity machine learning models such as deep neural networks have the ability to leverage subtle correlations and patterns to attain very low training error in learning tasks. As shown in Table \ref{note-length-dist} and Figure \ref{notes-length-box-plot}, there is a difference in our sample between mean length of patients who survived and those who had a negative outcome. A Mann-Whitney U test supplied further evidence, as we were able to reject the null hypothesis, i.e. distributions of the length of nursery notes are the same ($p=0.000$), in favor of our alternative hypothesis, i.e. notes for patients that do not survive are longer. \\

\begin{figure}[!h]
  
  \centering
    \includegraphics[width=0.8\textwidth]{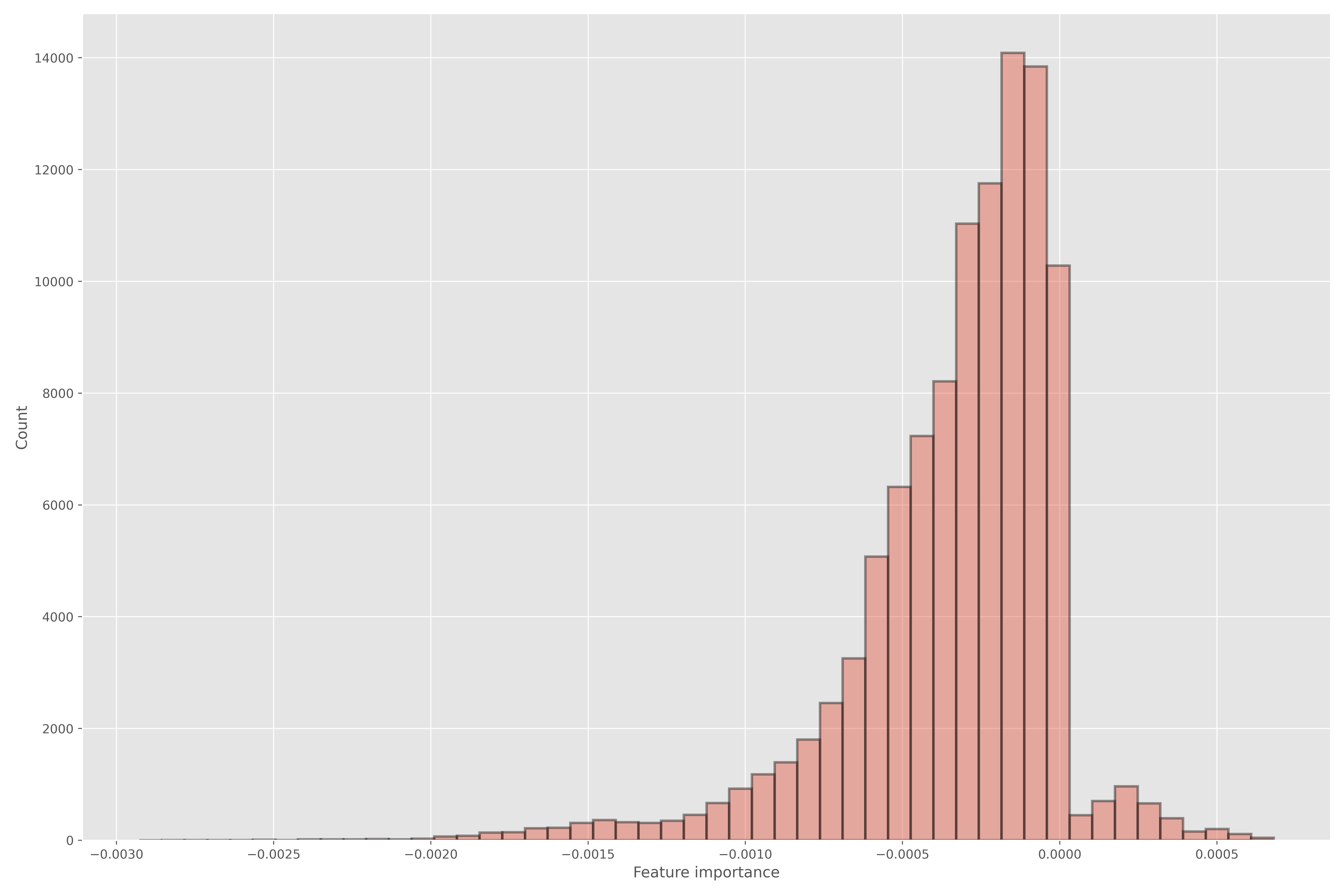}
    \caption{\label{shapley-hist} Distribution of approximate Shapley Values for padding characters. The histogram shows that most padding characters are deemed as evidence for survival by our model.}
\end{figure}

Having established that, we decided to investigate if our model was attending somehow to that difference in distributions. For this purpose we inspected the importance score of the padding characters used by our pre-processing pipeline, with most of them being regarded as evidence for survival, which is consistent with our original conjecture that the model considers that shorter notes are correlated with a survival outcome (shorter notes have more padding characters). Figure \ref{shapley-hist} shows the distribution of approximate Shapley Values for padding characters. 

\section{Discussion} Our convolutional model shows interesting performance on the MIMIC-III dataset, with consistent results across validation folds, showing evidence for good generalization. Validation ROC AUC (95\% CI [0.855689, 0.867888]) is competitive with published results in a comprehensive benchmark by \cite{PURUSHOTHAM2018} (95\% CI [0.873706, 0.882894] ROC AUC) and our previous work \cite{Caicedo-Torres2019} (95\% CI [0.870396, 0.876604] ROC AUC). Moreover, our model bypass some of the most important difficulties associated with the usage of physiological time series, i.e. inconsistent sampling times and missing values. On the other hand, the 24-hour version of our model still manages to surpass comfortably the SAPS-II baseline. \\

Our results are not directly comparable to those published by Grnarova el al \cite{Grnarova2016NeuralDE} given that we restricted our input window to the first 48 hours of patient stay, instead of using all available notes up until the time of discharge. Results published by Jo et al \cite{Jo2017CombiningLA} show their models performing under 0.84 ROC AUC for mortality prediction using MIMIC-III data (48 hour mark), which is well below our results here. On the other hand, the model Vital + EntityEmb reported by \cite{jin2018improving} uses physiological data and a substantial text preprocessing pipeline that involves a second neural network for Named Entity Recognition.Table \ref{auccomparison} shows reported performance results for relevant models, compared with the performance of our model.\\

\begin{table}[hbt!]

\centering

\resizebox{0.8\columnwidth}{!}{%
\begin{tabular}{l p{0.3\linewidth} r}
  \hline			
   Model & Type & ROC AUC \\
   \hline
   Physiological models & &  \\	
   \hline
   GRU-D \cite{DBLP:journals/corr/ChePCSL16} & Recurrent  & 0.8527 $\pm $ 0.003 \\
   MMDL \cite{PURUSHOTHAM2018} & Hybrid  & 0.8783 $\pm $ 0.0037 \\
   ISeeU & ConvNet & 0.8735 $\pm$ 0.0025\\
   \hline
   NLP models & &   \\	
   \hline
   LSTM+E+T+D \cite{Jo2017CombiningLA} & Recurrent & $<$ 0.84 \\
   Vital + EntityEmb \cite{jin2018improving} & Hybrid (text \& physiological inputs) & 0.8734 $\pm$ 0.0019 \\
   ISeeU2 (our work) & ConvNet & 0.8629 $\pm$ 0.0058  \\
   
  \hline  
\end{tabular}%

}
\caption{\label{auccomparison} Our results and results reported by related works. ROC AUC results are mean and standard deviation from a 5-fold cross validation run, except LSTM+E+T+D and Vital + EntityEmb, which report a single result over the test set. }
\end{table}

By using text notes as input we are using not the raw physiological data but healthcare workers perceptions and judgement in the form of free-text notes, giving us access to higher level concepts not present in said physiological data. On the other hand, MIMIC-III notes are very noisy, with frequent misspellings, typos and a lack of standardized naming (i.e. writing \textit{vancomycin} vs \textit{vanc}), which makes them not an optimal learning substrate. However, we have been able to show that deep learning models are able to separate useful signals from such noise, by keeping our pre-processing pipeline very basic. Another interesting take on the usage of free text notes is that deep models can leverage meta-data such as note length, as our evidence suggest. This phenomena is comparable to observations made in \cite{pmlr-v56-Lipton16, RazavianS15} regarding the ability of deep neural networks to exploit patterns of missingness in physiological patient data to attain better predictive performance: Certain physiological measurements are taken more or less frequently according to the state of the patient, providing additional and useful metadata. This kind of flexibility and power is outside the reach of more traditional statistical modeling techniques as the ones behind risk scores as SAPS-II. \\

Our visualization approach allows to easily locate the parts of the notes the deep learning model is attending to, which can then be compared to clinical expectations. In this way the potential users can be certain that the reasons behind the predictions are sound and align properly with medical knowledge, as opposed to being evidence of statistical artifacts being leveraged by the model. It is interesting to note that our results suggest our model and text heatmap visualization could be used to annotate medical notes for, at some point, easier handling by ICU staff. Finally, it's worth noticing that our usage of Shapley Values was instrumental to discover how the network regarded the padding introduced in shorter nursing notes.

\section{Limitations and future work}
Limitations of our study include the fact that we do not have access to some pre-admission data, and that we are using a retrospective, single center cohort. Also given the moderate size of our dataset we are only reporting cross-validation results without a proper test set result. An additional limitation is that high-quality nursing notes may not be available for a substantial number of patients in other critical care settings, which could hurt the performance of our model. Finally, the common misspellings and other noise present in the medical notes may affect the quality of the explanations, giving rise to counterintuitive results.\\

In future work we intend to investigate the usage of a more robust pre-processing pipeline, and assess whether there is any performance improvement attributable to its usage. Also we intend to evaluate how our approach fares in a situation where limited-quality notes are the only training data available. Finally we plan to explore the joint usage of physiological time series data and free-text medical notes to train a multi-modal deep learning model and compare its performance with our current approach. \\

\section{Conclusion}

In this paper we have presented ISeeU2, a convolutional neural network for the prediction of mortality using free-text nursing notes from MIMIC-III. We showed that our model is able to offer performance competitive with that of much more complex models with little text pre-processing, while at the same time providing visual explanations of feature importance based on coalitional game theory that allow users to gain insight on the reasons behind predicted outcomes. Our visualizations also provide a way to annotate free-text medical notes with markers to flag parts correlated with predictions of survival and death. We have also shown that nursing notes could be rich enough to capture the concepts needed for mortality prediction at a level of accuracy far higher than what is currently possible with traditional statistical techniques.

\section{Acknowledgments}
The authors thank Dr. Janet Liang, PhD (MBChB, FANZCA, FCICM), for her invaluable help and expertise. GPU access and computing services provided by the Data Centre at the Service and Cloud Computing Research Lab. Hosting services managed by Bumjun Kim, Senior Technician \href{mailto:bumjun.kim@aut.ac.nz}{bumjun.kim@aut.ac.nz} and ICT. The Data Centre is part of the School of Engineering, Computer and Mathematical Sciences, Auckland University of Technology.



\section*{References}
\bibliographystyle{elsarticle-num} 
\bibliography{phd}

\begin{thebibliography}{10}
\expandafter\ifx\csname url\endcsname\relax
  \def\url#1{\texttt{#1}}\fi
\expandafter\ifx\csname urlprefix\endcsname\relax\def\urlprefix{URL }\fi
\expandafter\ifx\csname href\endcsname\relax
  \def\href#1#2{#2} \def\path#1{#1}\fi

\bibitem{Grasselli2020}
G.~Grasselli, A.~Pesenti, M.~Cecconi, {Critical Care Utilization for the
  COVID-19 Outbreak in Lombardy, Italy}, JAMA (2020).
\newblock \href {https://doi.org/10.1001/jama.2020.4031}
  {\path{doi:10.1001/jama.2020.4031}}.

\bibitem{Emanuel2020}
E.~J. Emanuel, G.~Persad, R.~Upshur, B.~Thome, M.~Parker, A.~Glickman,
  C.~Zhang, C.~Boyle, M.~Smith, J.~P. Phillips, {Fair Allocation of Scarce
  Medical Resources in the Time of Covid-19}, New England Journal of Medicine
  (2020).
\newblock \href {https://doi.org/10.1056/nejmsb2005114}
  {\path{doi:10.1056/nejmsb2005114}}.

\bibitem{Rapsang:2014aa}
A.~G. Rapsang, D.~C. Shyam,
  \href{http://www.ncbi.nlm.nih.gov/pmc/articles/PMC4033855/}{{Scoring systems
  in the intensive care unit: A compendium}}, Indian Journal of Critical Care
  Medicine : Peer-reviewed, Official Publication of Indian Society of Critical
  Care Medicine 18~(4) (2014) 220--228.
\newblock \href {https://doi.org/10.4103/0972-5229.130573}
  {\path{doi:10.4103/0972-5229.130573}}.
\newline\urlprefix\url{http://www.ncbi.nlm.nih.gov/pmc/articles/PMC4033855/}

\bibitem{PURUSHOTHAM2018}
S.~Purushotham, C.~Meng, Z.~Che, Y.~Liu,
  \href{http://www.sciencedirect.com/science/article/pii/S1532046418300716}{{Benchmarking
  Deep Learning Models on Large Healthcare Datasets}}, Journal of Biomedical
  Informatics (2018).
\newblock \href {https://doi.org/https://doi.org/10.1016/j.jbi.2018.04.007}
  {\path{doi:https://doi.org/10.1016/j.jbi.2018.04.007}}.
\newline\urlprefix\url{http://www.sciencedirect.com/science/article/pii/S1532046418300716}

\bibitem{Caicedo-Torres2019}
W.~Caicedo-Torres, J.~Gutierrez,
  \href{https://www.sciencedirect.com/science/article/pii/S1532046419301881?dgcid=author}{{ISeeU:
  Visually interpretable deep learning for mortality prediction inside the
  ICU}}, Journal of Biomedical Informatics 98 (2019) 103269.
\newblock \href {https://doi.org/10.1016/J.JBI.2019.103269}
  {\path{doi:10.1016/J.JBI.2019.103269}}.
\newline\urlprefix\url{https://www.sciencedirect.com/science/article/pii/S1532046419301881?dgcid=author}

\bibitem{annurev-bioeng-071516-044442}
D.~Shen, G.~Wu, H.-I. Suk,
  \href{http://dx.doi.org/10.1146/annurev-bioeng-071516-044442}{{Deep Learning
  in Medical Image Analysis}}, Annual Review of Biomedical Engineering 19~(1)
  (2017) null.
\newblock \href {https://doi.org/10.1146/annurev-bioeng-071516-044442}
  {\path{doi:10.1146/annurev-bioeng-071516-044442}}.
\newline\urlprefix\url{http://dx.doi.org/10.1146/annurev-bioeng-071516-044442}

\bibitem{Shickel2018}
B.~Shickel, P.~J. Tighe, A.~Bihorac, P.~Rashidi, {Deep EHR: A Survey of Recent
  Advances in Deep Learning Techniques for Electronic Health Record (EHR)
  Analysis.}, IEEE journal of biomedical and health informatics 22~(5) (2018)
  1589--1604.
\newblock \href {https://doi.org/10.1109/JBHI.2017.2767063}
  {\path{doi:10.1109/JBHI.2017.2767063}}.

\bibitem{Grnarova2016NeuralDE}
P.~Grnarova, F.~Schmidt, S.~L. Hyland, C.~Eickhoff, {Neural Document Embeddings
  for Intensive Care Patient Mortality Prediction}, CoRR abs/1612.0 (2016).

\bibitem{Cooper1997}
G.~F. Cooper, C.~F. Aliferis, R.~Ambrosino, J.~Aronis, B.~G. Buchanan,
  R.~Caruana, M.~J. Fine, C.~Glymour, G.~Gordon, B.~H. Hanusa, J.~E. Janosky,
  C.~Meek, T.~Mitchell, T.~Richardson, P.~Spirtes, {An evaluation of
  machine-learning methods for predicting pneumonia mortality}, Artificial
  Intelligence in Medicine (1997).
\newblock \href {https://doi.org/10.1016/S0933-3657(96)00367-3}
  {\path{doi:10.1016/S0933-3657(96)00367-3}}.

\bibitem{Johnson:2016aa}
A.~E.~W. Johnson, T.~J. Pollard, L.~Shen, L.-W.~H. Lehman, M.~Feng,
  M.~Ghassemi, B.~Moody, P.~Szolovits, L.~A. Celi, R.~G. Mark, {MIMIC-III, a
  freely accessible critical care database.}, Sci Data 3 (2016) 160035.
\newblock \href {https://doi.org/10.1038/sdata.2016.35}
  {\path{doi:10.1038/sdata.2016.35}}.

\bibitem{Jo2017CombiningLA}
Y.~Jo, L.~Lee, S.~Palaskar, {Combining LSTM and Latent Topic Modeling for
  Mortality Prediction}, ArXiv abs/1709.0 (2017).

\bibitem{Sushil2018}
M.~Sushil, S.~{\v{S}}uster, K.~Luyckx, W.~Daelemans, {Patient representation
  learning and interpretable evaluation using clinical notes}, Journal of
  Biomedical Informatics (2018).
\newblock \href {http://arxiv.org/abs/1807.01395} {\path{arXiv:1807.01395}},
  \href {https://doi.org/10.1016/j.jbi.2018.06.016}
  {\path{doi:10.1016/j.jbi.2018.06.016}}.

\bibitem{Si2019}
Y.~Si, K.~Roberts, {Deep Patient Representation of Clinical Notes via
  Multi-Task Learning for Mortality Prediction.}, AMIA Joint Summits on
  Translational Science proceedings. AMIA Joint Summits on Translational
  Science (2019).

\bibitem{jin2018improving}
M.~Jin, M.~T. Bahadori, A.~Colak, P.~Bhatia, B.~Celikkaya, R.~Bhakta,
  S.~Senthivel, M.~Khalilia, D.~Navarro, B.~Zhang, T.~Doman, A.~Ravi, M.~Liger,
  T.~Kass-hout, {Improving Hospital Mortality Prediction with Medical Named
  Entities and Multimodal Learning} (2018).
\newblock \href {http://arxiv.org/abs/1811.12276} {\path{arXiv:1811.12276}}.

\bibitem{LeCun:1998aa}
Y.~LeCun, L.~Bottou, Y.~Bengio, Haffner, {Gradient-Based Learning Applied to
  Document Recognition}, in: Proceedings of the IEEE, Vol.~86, 1998, pp.
  2278--2324.

\bibitem{Goodfellow-et-al-2016}
I.~Goodfellow, Y.~Bengio, A.~Courville, {Deep Learning}, MIT Press, 2016.

\bibitem{shapley:book1952}
L.~S. Shapley, {A Value for n-Person Games}, in: H.~W. Kuhn, A.~W. Tucker
  (Eds.), Contributions to the Theory of Games II, Princeton University Press,
  Princeton, 1953, pp. 307--317.

\bibitem{Strumbelj2010}
E.~Strumbelj, I.~Kononenko, S.~Wrobel, {An Efficient Explanation of Individual
  Classifications using Game Theory}, Journal of Machine Learning Research
  (2010).
\newblock \href {http://arxiv.org/abs/1606.05386} {\path{arXiv:1606.05386}},
  \href {https://doi.org/10.1145/2858036.2858529}
  {\path{doi:10.1145/2858036.2858529}}.

\bibitem{DBLP:journals/corr/ShrikumarGK17}
A.~Shrikumar, P.~Greenside, A.~Kundaje,
  \href{http://arxiv.org/abs/1704.02685}{{Learning Important Features Through
  Propagating Activation Differences}}, CoRR abs/1704.0 (2017).
\newblock \href {http://arxiv.org/abs/1704.02685} {\path{arXiv:1704.02685}}.
\newline\urlprefix\url{http://arxiv.org/abs/1704.02685}

\bibitem{Lundberg2017}
S.~M. Lundberg, S.~I. Lee, {A unified approach to interpreting model
  predictions}, in: Advances in Neural Information Processing Systems, 2017.
\newblock \href {http://arxiv.org/abs/1705.07874} {\path{arXiv:1705.07874}}.

\bibitem{SimonyanVZ13}
K.~Simonyan, A.~Vedaldi, A.~Zisserman,
  \href{http://arxiv.org/abs/1312.6034}{{Deep Inside Convolutional Networks:
  Visualising Image Classification Models and Saliency Maps}}, CoRR abs/1312.6
  (2013).
\newline\urlprefix\url{http://arxiv.org/abs/1312.6034}

\bibitem{DBLP:journals/corr/SpringenbergDBR14}
J.~T. Springenberg, A.~Dosovitskiy, T.~Brox, M.~A. Riedmiller,
  \href{http://arxiv.org/abs/1412.6806}{{Striving for Simplicity: The All
  Convolutional Net}}, CoRR abs/1412.6 (2014).
\newblock \href {http://arxiv.org/abs/1412.6806} {\path{arXiv:1412.6806}}.
\newline\urlprefix\url{http://arxiv.org/abs/1412.6806}

\bibitem{Abadi2015}
M.~Abadi, A.~Agarwal, P.~Barham, E.~Brevdo, Z.~Chen, C.~Citro, G.~Corrado,
  A.~Davis, J.~Dean, M.~Devin, S.~Ghemawat, I.~Goodfellow, A.~Harp, G.~Irving,
  M.~Isard, Y.~Jia, L.~Kaiser, M.~Kudlur, J.~Levenberg, D.~Man, R.~Monga,
  S.~Moore, D.~Murray, J.~Shlens, B.~Steiner, I.~Sutskever, P.~Tucker,
  V.~Vanhoucke, V.~Vasudevan, O.~Vinyals, P.~Warden, M.~Wicke, Y.~Yu, X.~Zheng,
  \href{http://download.tensorflow.org/paper/whitepaper2015.pdf}{{TensorFlow:
  Large-Scale Machine Learning on Heterogeneous Distributed Systems}}, None
  1~(212) (2015) 19.
\newblock \href {http://arxiv.org/abs/1603.04467} {\path{arXiv:1603.04467}},
  \href {https://doi.org/10.1038/nn.3331} {\path{doi:10.1038/nn.3331}}.
\newline\urlprefix\url{http://download.tensorflow.org/paper/whitepaper2015.pdf}

\bibitem{2014arXiv1412.6980K}
D.~Kingma, J.~Ba, {Adam: A Method for Stochastic Optimization}, ArXiv e-prints
  (dec 2014).
\newblock \href {http://arxiv.org/abs/1412.6980} {\path{arXiv:1412.6980}}.

\bibitem{Loper2002}
E.~Loper, S.~Bird, {NLTK}, 2002.
\newblock \href {https://doi.org/10.3115/1118108.1118117}
  {\path{doi:10.3115/1118108.1118117}}.

\bibitem{Gall1993}
J.~R. Gall, S.~Lemeshow, F.~Saulnier, {A New Simplified Acute Physiology Score
  (SAPS II) Based on a European/North American Multicenter Study}, JAMA: The
  Journal of the American Medical Association (1993).
\newblock \href {http://arxiv.org/abs/0402594v3} {\path{arXiv:0402594v3}},
  \href {https://doi.org/10.1001/jama.1993.03510240069035}
  {\path{doi:10.1001/jama.1993.03510240069035}}.

\bibitem{Johnson2018}
A.~E. Johnson, D.~J. Stone, L.~A. Celi, T.~J. Pollard, {The MIMIC Code
  Repository: Enabling reproducibility in critical care research}, Journal of
  the American Medical Informatics Association (2018).
\newblock \href {https://doi.org/10.1093/jamia/ocx084}
  {\path{doi:10.1093/jamia/ocx084}}.

\bibitem{Hochreiter:1997:LSM:1246443.1246450}
S.~Hochreiter, J.~Schmidhuber,
  \href{http://dx.doi.org/10.1162/neco.1997.9.8.1735}{{Long Short-Term
  Memory}}, Neural Comput. 9~(8) (1997) 1735--1780.
\newblock \href {https://doi.org/10.1162/neco.1997.9.8.1735}
  {\path{doi:10.1162/neco.1997.9.8.1735}}.
\newline\urlprefix\url{http://dx.doi.org/10.1162/neco.1997.9.8.1735}

\bibitem{DBLP:journals/corr/Lipton16a}
Z.~C. Lipton, \href{http://arxiv.org/abs/1606.03490}{{The Mythos of Model
  Interpretability}}, ICML Workshop on Human Interpretability in Machine
  Learning abs/1606.0 (2016) 96--100.
\newblock \href {http://arxiv.org/abs/arXiv:1606.03490v1}
  {\path{arXiv:arXiv:1606.03490v1}}.
\newline\urlprefix\url{http://arxiv.org/abs/1606.03490}

\bibitem{DBLP:journals/corr/ChePCSL16}
Z.~Che, S.~Purushotham, K.~Cho, D.~Sontag, Y.~Liu,
  \href{http://arxiv.org/abs/1606.01865}{{Recurrent Neural Networks for
  Multivariate Time Series with Missing Values}}, CoRR abs/1606.0 (2016).
\newline\urlprefix\url{http://arxiv.org/abs/1606.01865}

\bibitem{pmlr-v56-Lipton16}
Z.~C. Lipton, D.~Kale, R.~Wetzel,
  \href{http://proceedings.mlr.press/v56/Lipton16.html}{{Directly Modeling
  Missing Data in Sequences with RNNs: Improved Classification of Clinical Time
  Series}}, in: F.~Doshi-Velez, J.~Fackler, D.~Kale, B.~Wallace, J.~Weins
  (Eds.), Proceedings of the 1st Machine Learning for Healthcare Conference,
  Vol.~56 of Proceedings of Machine Learning Research, PMLR, Northeastern
  University, Boston, MA, USA, 2016, pp. 253--270.
\newline\urlprefix\url{http://proceedings.mlr.press/v56/Lipton16.html}

\bibitem{RazavianS15}
N.~Razavian, D.~Sontag, \href{http://arxiv.org/abs/1511.07938}{{Temporal
  convolutional neural networks for diagnosis lab tests}}, 25 November
  abs/1511.0 (2015) 1--17.
\newblock \href {http://arxiv.org/abs/1151.07938v1}
  {\path{arXiv:1151.07938v1}}, \href
  {https://doi.org/10.1051/0004-6361/201527329}
  {\path{doi:10.1051/0004-6361/201527329}}.
\newline\urlprefix\url{http://arxiv.org/abs/1511.07938}

\end{thebibliography}




%
\end{document}